\documentclass[letterpaper,10pt,conference]{ieeeconf}

\IEEEoverridecommandlockouts
\usepackage{amsmath}
\usepackage{amssymb}
\usepackage{booktabs}
\usepackage{graphicx}
\usepackage{makecell}
\usepackage{subcaption}
\usepackage{tikz}
\usepackage{tikz-3dplot}
\usepackage{url}

\usetikzlibrary{
    arrows.meta,
    calc,
    positioning
}

\begin{document}

\title{\LARGE\bfseries
Beyond Shallow-Water Photorealism:\\
Physically and Sensor-Grounded Simulation\\
for Deep-Sea Robotics
}

\author{
Michele Grimaldi$^{1}$,
Enrico Di Maria$^{2}$,
Ignacio Carlucho$^{1}$
and Yvan R. Petillot$^{1}$%
\thanks{$^{1}$School of Engineering \& Physical Sciences,
Heriot-Watt University, Edinburgh, UK.
{\tt\small m.grimaldi@hw.ac.uk}}%
\thanks{$^{2}$Engineering Department,
Japan Agency for Marine-Earth Science and Technology,
Yokosuka, Japan.}%
}

\maketitle
\thispagestyle{empty}
\pagestyle{empty}

\begin{abstract}
Many recent underwater simulators emphasize visual realism at the expense of physical fidelity, focusing on shallow-water effects with limited relevance in deep-water environments and high computational cost. In this work, we shift the focus toward deep-sea physical and sensor realism. We present a physics- and sensor-grounded extension of the Stonefish simulator that augments its hydrodynamic models with stochastic IMU and DVL drift, magnetometer disturbances, higher-order hydrodynamics, terramechanics, pressure-driven environmental variability, and physically based underwater optics. These additions are designed to better capture the forces and measurements shaping the behavior of deep-ocean AUVs, ROVs, landers, ASVs, and gliders, while remaining compatible with real-time simulation. 
This work advances underwater simulation toward more representative deep-sea operating conditions, which is particularly relevant for long-duration navigation and learning-based autonomy, where inaccurate sensor and environmental models introduce non-physical artifacts and overly optimistic performance. While challenges remain, including complex fluid--structure interactions and full environmental stochasticity, the proposed framework provides a practical foundation for navigation, perception, and autonomy research under deep-sea conditions.
\end{abstract}

\begingroup
\renewcommand\thefootnote{}
\endgroup


\section{Introduction}

Simulation is central to underwater robotics, enabling rapid development, reproducible testing, and safe evaluation of perception, navigation, and control algorithms. However, most simulators target shallow-water AUVs and ROVs, offering limited support for crawlers, landers, or deep-sea variants. This gap is critical, as deep-sea missions are among the most costly and demanding in marine robotics, where limited access and high deployment costs make high-fidelity simulation essential for validation and mission readiness. These simulators often inherit priorities from computer graphics and game engines, emphasizing photorealistic visuals under the assumption that visual fidelity implies physical realism, despite effects such as caustics, crepuscular rays, and cinematic scattering having limited impact in deep water or being computationally prohibitive for real-time autonomy. As missions increasingly operate in deep water, perception and state estimation are dominated by spectral attenuation, turbidity, low signal-to-noise ratios, sensor distortions, and navigation constraints arising from long-horizon inertial drift and intermittent Doppler velocity measurements in GPS-denied conditions, requiring physics-based modeling rather than enhanced visual aesthetics.

\begin{figure}[h]
    \centering
    \includegraphics[width=0.91\linewidth]{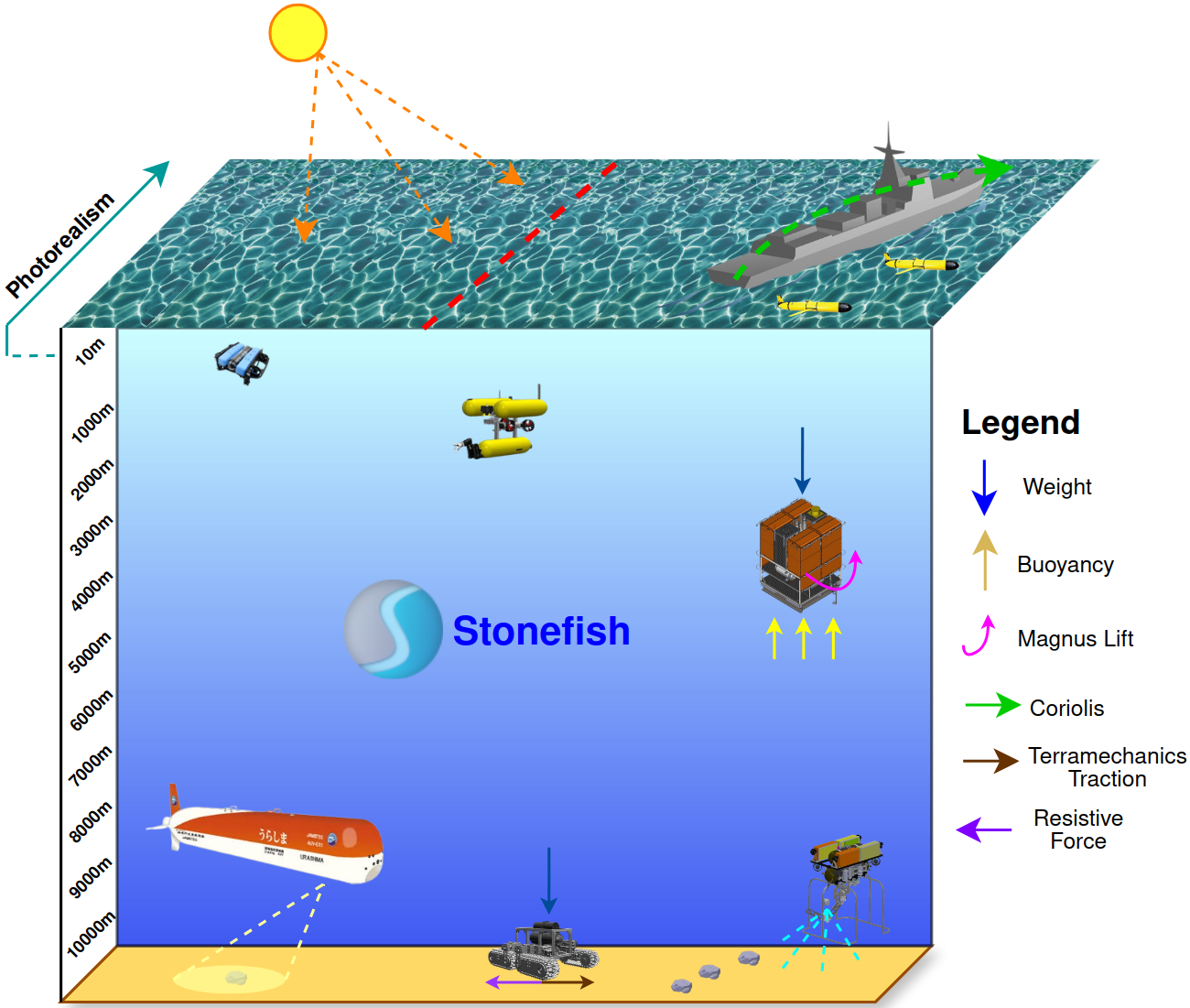}
    \caption{Overview of the extended Stonefish simulator, incorporating advanced hydrodynamics, terramechanics, and lift, Coriolis, and Magnus effects for realistic simulation of diverse underwater vehicles.}
    \label{fig:placeholder}
    \vspace{-1em}
\end{figure} 

This aligns with broader observations in robotics that simulation is most effective when grounded in physically accurate models of sensing, actuation, and environmental interaction rather than visual realism alone \cite{role_of}. Most widely used platforms, including Unity, Unreal, or custom GPU-based simulators, provide limited support for these conditions, often omitting realistic inertial and DVL sensing, magnetic disturbances, higher-order hydrodynamics, and benthic interactions. Real-time rigid-body solvers such as Bullet are efficient and stable but may not fully capture the number and coupling of forces required for deep-sea realism, while offline CFD or FEM approaches remain impractical for closed-loop autonomy or large-scale experimentation \cite{app13020680}. These limitations are amplified in learning-based autonomy, where model-free reinforcement learning methods may exploit non-physical artifacts when sensor and environmental models are inaccurate \cite{712192, tu2019gapmodelbasedmodelfreemethods, osti_2282016}. Inaccurate inertial or velocity sensing can mask estimator drift and bias accumulation, leading to over-optimistic navigation and control performance. Physics-based simulation is therefore a prerequisite for reliable perception and control. Previous version of Stonefish~\cite{grimaldi2025stonefishsupportingmachinelearning} provided a robust, open-source foundation emphasizing physical modeling over visual fidelity but lacks several key components for realistic deep-sea robotics. We present a physics- and sensor-grounded extension of Stonefish designed to address these gaps through an integrated optics, physics, and sensing-centric real-time simulation framework. The system combines these components within a ROS-compatible architecture, enabling long-horizon navigation, contact-rich interaction, and learning-based autonomy studies under deep-sea conditions. Figure~\ref{fig:placeholder} illustrates some of the main simulator extensions. Our main contributions are:
\begin{itemize}
\item A unified stochastic sensor framework for IMU, DVL, and magnetometer sensing, including physically motivated drift, per-beam DVL simulation, and hard- and soft-iron magnetic disturbances.
\item A surface-based hydrodynamic and environmental interaction model capturing lift, Magnus, and Coriolis effects, pycnocline-consistent density variation, depth-dependent hydrodynamic coefficients.
\item A lightweight Bekker-based terramechanics model for seabed locomotion, which now enables the deployment of seabed crawlers in Stonefish.
\item An underwater camera and rendering pipeline with configurable intrinsics, parametric lens distortion and deep-sea optical effects.
\end{itemize}


\begin{table*}[htbp]
\centering
\scriptsize
\renewcommand{\arraystretch}{}
\caption{IMU and DVL sensor modeling in underwater simulators.}
\label{tab:sensor_comparison_concise_updated}
\begin{tabular}{|l|c|p{6cm}|c|p{7cm}|}
\hline
\textbf{Simulator} & \textbf{IMU} & \textbf{IMU Model} & \textbf{DVL} & \textbf{DVL Model} \\
\hline
\makecell{\textbf{Stonefish} (Ours)} & Yes &
Physics-based MEMS IMU with scale/misalignment errors, bias random walk, Gauss--Markov noise, temperature drift, and vibration effects. &
Yes &
Physically based 4-beam DVL with bottom- and water-track, per-beam velocities, noise, and stochastic drift via bias random walk. \\
\hline
\textbf{HoloOcean} & Yes &
Ideal IMU with configurable Gaussian noise and bias; returns angular velocity and linear acceleration. &
Yes &
Acoustic DVL simulating Doppler-based velocity relative to seabed or objects, with configurable noise and range limits. \\
\hline
\textbf{OceanSim} & No &
\centering N\/A &
Yes &
4-beam Janus DVL with adaptive update rate, beam dropouts, and noisy velocity returns for realistic navigation testing. \\
\hline
\textbf{DAVE} & No &
\centering N\/A &
Yes &
Gazebo-based DVL plugin: velocity from ground truth plus Gaussian noise; altitude from 4 RaySensors modeling sonar beams. \\
\hline
\end{tabular}
\vspace{-0.25cm}
\end{table*}

\section{Related Work}
Simulation is a core enabler of underwater robotics research, supporting the development and evaluation of perception, navigation, and control algorithms prior to costly and risky field deployment. Despite its importance, the landscape of underwater simulation remains fragmented and often emphasizes shallow-water scenarios. General-purpose frameworks such as DAVE provide ROS-integrated aquatic simulation environments for vehicles and manipulators, but typically rely on simplified environmental interactions and sensing models that limit their applicability in deep-sea conditions~\cite{zhang2022daveaquaticvirtualenvironment}. At the same time, a growing number of simulators built on game engines or GPU-accelerated rendering pipelines emphasize visually rich underwater scenes, supporting optical and acoustic sensors with high-quality rendering~\cite{10638434, amer2023unavsimvisuallyrealisticunderwater}. While highly effective for visualization and shallow-water perception studies, these platforms often focus on visual realism over fully physics-based modeling, providing limited support for deep-water optical degradation, sensor drift, or long-term environmental dynamics. More recent GPU-based efforts, such as OceanSim, aim to support efficient underwater perception with physics-inspired rendering for a limited set of sensors, but still typically offer restricted modeling of vehicle dynamics, hydrodynamic forces, and long-term sensor behavior~\cite{song2025oceansimgpuacceleratedunderwaterrobot}. Similarly,  MarineGym~\cite{chu2025marinegymhighperformancereinforcementlearning} emphasizes high-performance reinforcement learning, prioritizing training efficiency over sensing fidelity and detailed environmental modeling.
Surveys of underwater robotics simulators consistently highlight trade-offs among physics fidelity, sensor realism, and computational cost, with deep-sea environments remaining less represented in existing platforms~\cite{aldhaheri2025underwaterroboticsimulatorsreview}. Stonefish distinguishes itself by emphasizing rigid-body and hydrodynamic modeling derived from vehicle geometry, coupled vehicle–manipulator dynamics, and tight ROS integration~\cite{ grimaldi2025stonefishsupportingmachinelearning}. However, like many existing simulators, Stonefish provides limited support for deep-sea–specific sensing and environmental effects, including inertial and magnetic drift, pressure-driven environmental variation, and physically grounded deep-water optical attenuation. These limitations are increasingly problematic as learning-based methods are adopted for underwater autonomy, since data-driven machine learning and reinforcement learning approaches depend critically on the physical and sensor fidelity of the simulation environment. To address these limitations, we introduce a physics- and sensor-centric extension to the Stonefish framework designed to support deep-sea robotic operations, with a particular focus on physically grounded sensing and environmental interactions.

\begin{figure*}
\scriptsize
\centering
\begin{tikzpicture}[
    node distance=1.cm, 
    auto, 
    thick, 
    every node/.style={scale=0.85, draw, rectangle, align=center, inner sep=5pt}
]
    \node[fill=blue!10] (true) {True Motion: \\ $\boldsymbol{\omega}_\text{true}, \mathbf{a}_\text{true}$};
    
    \node[fill=green!10, right of=true, xshift=2.5cm] (misalign) {Scale Factor \& Misalignment: \\ $\mathbf{M}(\mathbf{1}+\mathbf{S})$};
    
    \node[fill=red!10, right of=misalign, xshift=3.5cm] (noise) {Gauss--Markov/White Noise: \\ $\mathbf{n}^\text{GM}(t), \boldsymbol{\eta}$};

     \node[fill=orange!10, below of=noise, yshift=-0.5cm] (bias) {Bias Random Walk: \\ $\mathbf{b}(t)$};
    
    \node[fill=purple!10, right of=noise, xshift=3.5cm] (vib) {Vibration Effects: \\ $\mathbf{A}_v \sin(2\pi f_v t)$};

    \node[fill=yellow!10, below of=vib, yshift=-0.5cm] (temp) {Temperature Drift: \\ $(T-T_\text{ref})\mathbf{k}_T$};
    
    \node[fill=gray!10, right of=vib, xshift=2.5cm] (meas) {Measured Output: \\ $\boldsymbol{\omega}_s, \mathbf{a}_s$};

    \draw[->] (true) -- (misalign);
    \draw[->] (misalign) -- (noise);
    \draw[->] (noise) -- (vib);
    \draw[->] (vib) -- (meas);

    \draw[->] (bias.north) -- (noise.south); 
    \draw[->] (temp.north) -- (vib.south);
\end{tikzpicture}
\caption{IMU signal processing flow. True motion is affected by scale/misalignment, bias, temporally correlated noise, temperature drift, and vibration before producing the measured outputs.}
\label{fig:imu_signal_flow_fixed}
\vspace{-0.5cm}
\end{figure*}
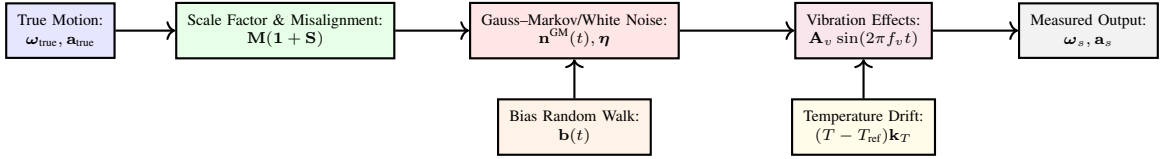

\section{IMU Noise and Bias Modeling}

Accurate inertial sensing is essential for long-duration underwater navigation, where modelling errors accumulate as drift. As shown in Table~\ref{tab:sensor_comparison_concise_updated}, most simulators use simplified white-noise and static-bias IMU models that fail to reproduce the long-term stochastic behaviour of real MEMS devices.
We implement a unified stochastic IMU model, illustrated in Fig.~\ref{fig:imu_signal_flow_fixed}, that captures short-term noise and long-term bias dynamics while remaining compatible with real-time simulation. The proposed model incorporates the following effects:

\begin{enumerate}
    \item \textbf{Scale Factor and Misalignment:} Each axis includes a small scale error and cross-axis misalignment. The true angular velocity $\boldsymbol{\omega}_{\text{true}}$ and linear acceleration $\mathbf{a}_{\text{true}}$ are transformed as
    \begin{equation}
        \boldsymbol{\omega}_s = \mathbf{M} (\mathbf{I} + \mathbf{S}) \boldsymbol{\omega}_{\text{true}}, \quad
        \mathbf{a}_s = \mathbf{M} (\mathbf{I} + \mathbf{S}) \mathbf{a}_{\text{true}},
    \end{equation}
    where $\mathbf{M}$ represents the misalignment matrix and $\mathbf{S}$ the diagonal scale factors.

    \item \textbf{Stochastic Bias Drift:} Slowly varying gyroscope and accelerometer biases $\mathbf{b}_\omega, \mathbf{b}_a$ are modeled as Brownian motion,
    \begin{equation}
        \mathbf{b}(t+\Delta t) = \mathbf{b}(t) + \boldsymbol{\eta} \sqrt{\Delta t},
    \end{equation}
    where $\boldsymbol{\eta} \sim \mathcal{N}(0, \sigma_{\text{bias}}^2)$ governs long-term drift accumulation.

    \item \textbf{Temporally Correlated Noise:} To reproduce colored noise observed in MEMS sensors, a first-order Gauss--Markov process is applied,
    \begin{equation}
        \mathbf{n}^{\text{GM}}(t+\Delta t) = \phi \mathbf{n}^{\text{GM}}(t) + \sqrt{1-\phi^2}\,\mathbf{w}(t),
    \end{equation}
    where $\phi$ is the correlation coefficient and $\mathbf{w}(t) \sim \mathcal{N}(0, \sigma^2)$.

    \item \textbf{Quantization and Range Limits:} Measurements are quantized according to the sensor’s least significant bit (LSB) and clipped to its physical operating range.

    \item \textbf{Temperature-Dependent Drift:} A linear temperature sensitivity term models thermal bias variation,
    \begin{equation}
        \mathbf{a}_{\text{meas}} \leftarrow \mathbf{a}_{\text{meas}} + (T - T_\text{ref}) \mathbf{k}_T,
    \end{equation}
    where $T_\text{ref}$ is the reference temperature and $\mathbf{k}_T$ the temperature coefficient.

    \item \textbf{Vibration-Induced Noise:} Low-amplitude sinusoidal components are added to approximate high-frequency vibration effects commonly induced by thrusters and mechanical coupling.
\end{enumerate}

The final measured signals are obtained by combining the above effects with additive white noise:
\begin{equation}
\boldsymbol{\omega}_{\text{meas}} = \boldsymbol{\omega}_s + \mathbf{b}_\omega + \mathbf{n}^{\text{GM}}_\omega + \mathbf{n}_\omega, \quad
\mathbf{a}_{\text{meas}} = \mathbf{a}_s + \mathbf{b}_a + \mathbf{n}^{\text{GM}}_a + \mathbf{n}_a.
\end{equation}

All noise and drift parameters are fully parametrizable and can be tuned to match target sensor specifications.

\begin{figure}[t]
\centering
\includegraphics[width=0.49\textwidth]{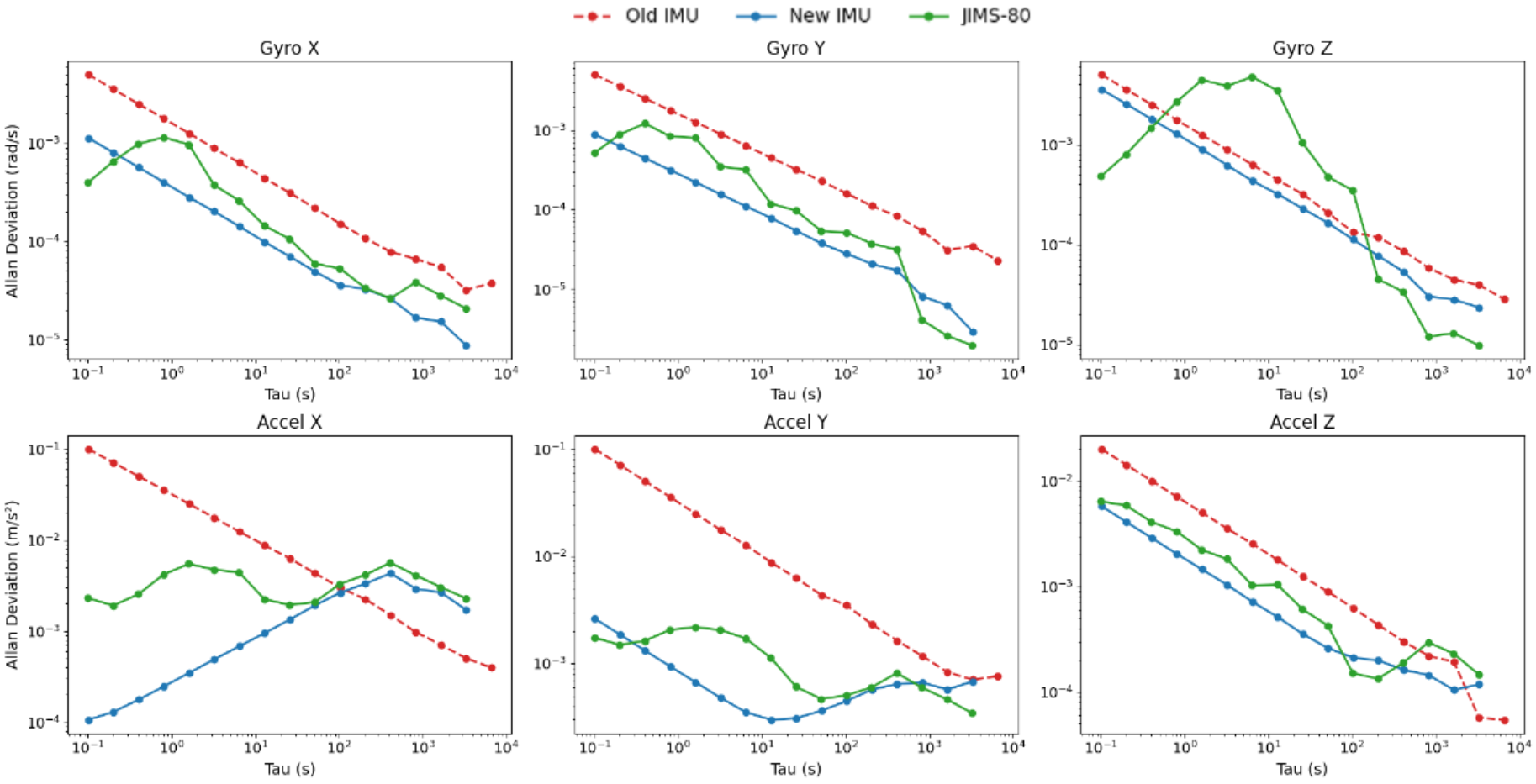}
\caption{Allan deviation comparison of the old IMU (red, dashed), new IMU (blue, solid), and JIMS-80 (green, solid). The x-axis represents averaging time $\tau$ in seconds, and the y-axis shows the Allan deviation of the measurements. Top row corresponds to gyroscope (rad/s) and bottom row to accelerometer (m/s²).}
\label{fig:allan}
\vspace{-0.5cm}
\end{figure}

\subsection{Allan Variance Test}

To evaluate the long-horizon stochastic behavior of the simulated IMU, we performed an Allan variance analysis~\cite{5570702} using over 9~h of data from a real JIMS-80 IMU, the original Stonefish IMU model, and the new model. For a time series $x(t)$ sampled at $\tau_0$, the Allan variance is defined as
\begin{equation}
    \sigma^2(\tau) = \frac{1}{2 (N-1)} \sum_{k=1}^{N-1} \left( \bar{x}_{k+1} - \bar{x}_k \right)^2,
\end{equation}
where $\bar{x}_k$ is the mean over the $k$-th cluster of duration $\tau$ and $N$ is the number of clusters. Allan variance enables separation of short-term noise (angular random walk, ARW) from slow bias drift (bias instability, BI) and long-term rate random walk (RRW). Figure~\ref{fig:allan} compares Allan deviation curves for the three datasets. The original IMU model (red, dashed) is dominated by white noise, with gyroscope ARW of approximately $1.6\times10^{-3}$~rad/s and negligible bias instability ($\sim3\times10^{-5}$~rad/s), which is not representative of typical MEMS sensors. In contrast, the proposed IMU model (blue, solid) exhibits both reduced ARW ($\sim5\times10^{-4}$~rad/s) and clear bias instability in the $10^{-5}$--$10^{-6}$~rad/s range, closely matching the stochastic structure of the real JIMS-80 IMU (green, solid). These results demonstrate that the proposed model reproduces both short-term stochastic noise and long-term bias drift that are absent in the original simulator, yielding realistic long-horizon inertial behavior. Since all noise and drift parameters are fully tunable, the model can be calibrated to match target IMU specifications; the values shown here are illustrative and serve to validate the modeling approach.

\section{DVL Drift and Bias}
\begin{figure}[t]
    \centering
    \includegraphics[width=\linewidth]{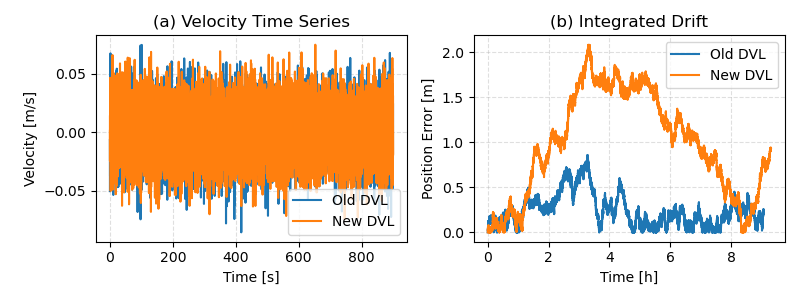}
    \caption{Stationary DVL validation over 3 hours: zero-velocity time series (left) and dead-reckoning drift (right).}
    \label{fig:dvl_val}
    \vspace{-1em}
\end{figure}

The Doppler Velocity Log (DVL) simulation is extended to provide realistic three-axis velocity measurements relative to both the seabed and water currents, including per-beam velocity outputs. Unlike the original Stonefish implementation, which applied an ad hoc drift-rate parameter, the proposed model introduces a physically motivated, slowly varying bias that produces natural long-horizon drift. Let $\mathbf{v}_\text{true}$ denote the vehicle velocity relative to the seabed. Bottom-track measurements are obtained by projecting $\mathbf{v}_\text{true}$ along four acoustic beams $\mathbf{d}_i$ tilted by $\theta$ from the vertical,
\begin{equation}
    v_{\text{beam},i} = \mathbf{v}_\text{true} \cdot \mathbf{d}_i + \delta v_i ,
\end{equation}
where $\delta v_i$ represents additive beam noise. For water-track operation, the measured velocity accounts for the sampled water velocity $\mathbf{v}_\text{water}$,
\begin{equation}
    \mathbf{v}_\text{meas} = \mathbf{v}_\text{true} - \mathbf{v}_\text{water} + \boldsymbol{\eta},
\end{equation}
with $\boldsymbol{\eta}$ denoting short-term measurement noise.
To reproduce realistic drift behavior, a slowly varying bias vector $\mathbf{b}(t)$ is introduced:
\begin{equation}
    \mathbf{b}(t+\Delta t) = \mathbf{b}(t) + \boldsymbol{\xi}\,\Delta t ,
\end{equation}
where $\boldsymbol{\xi}$ is a small zero-mean random increment. The resulting velocity measurement is
\begin{equation}
    \mathbf{v}_\text{DVL} = \mathbf{v}_\text{true} + \mathbf{b}(t) + \boldsymbol{\eta},
\end{equation}
and per-beam outputs are generated as
\begin{equation}
    v_{\text{beam},i}^\text{DVL} = (\mathbf{v}_\text{true} + \mathbf{b}(t)) \cdot \mathbf{d}_i + \eta_i ,
\end{equation}
ensuring consistency between aggregate velocity and beam-level Doppler measurements. Quantitative validation is summarized in Fig.~\ref{fig:dvl_val}. Both models yield near-zero mean velocities ($7.0\times10^{-6}$~m/s for the old DVL and $2.8\times10^{-5}$~m/s for the new DVL), which are negligible relative to the measurement noise, and exhibit comparable short-term noise levels (standard deviation $\approx 0.02$~m/s). However, over a 3-hour horizon the original white-noise DVL model exhibits artificial drift cancellation and unrealistically low error growth ($0.21$~m/h), whereas the proposed bias-driven model accumulates substantially larger and monotonic drift ($0.56$~m/h), consistent with bias-driven integration effects observed in real DVL systems.

\section{Magnetometer Disturbance Modeling}

\begin{figure}[t]
    \centering
    \includegraphics[width=0.49\linewidth]{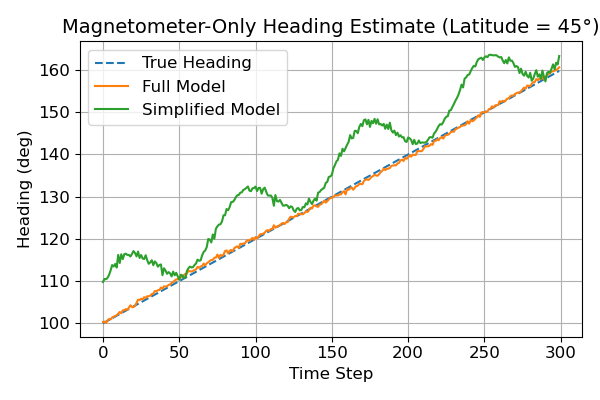}
    \includegraphics[width=0.49\linewidth]{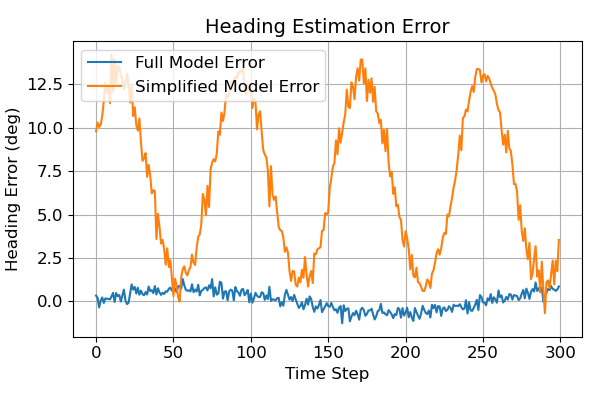}
    \caption{Magnetometer-only heading estimation at 45° latitude. Left: heading estimates. Right: heading error.} 
    \label{fig:magnetometer_heading}
    \vspace{-0.5cm}
\end{figure}

We implement a physically grounded magnetometer model that produces three-axis magnetic field measurements in the sensor frame while accounting for geomagnetic variation, vehicle- and infrastructure-induced disturbances, and sensor imperfections. In contrast, HoloOcean provides a fixed global reference vector in the sensor frame with optional noise, without modeling latitude-dependent geomagnetic variation or hard- and soft-iron effects; OceanSim and DAVE do not include a magnetometer sensor. The Earth’s magnetic field is obtained from the World Magnetic Model (WMM) or the International Geomagnetic Reference Field (IGRF) as a function of latitude, longitude, altitude, and time:
\begin{equation}
    \mathbf{B}_\text{earth} = \mathrm{WMM}(\text{lat}, \text{lon}, h, t).
\end{equation}
Hard-iron disturbances from nearby ferromagnetic objects are modeled as magnetic dipoles $\mathbf{B}_\text{HI}$, while soft-iron effects are represented by a linear distortion matrix $\mathbf{S}$:
\begin{equation}
    \mathbf{B}_\text{dist} = \mathbf{S} \left( \mathbf{B}_\text{earth} + \mathbf{B}_\text{HI} \right).
\end{equation}
Zero-mean noise is added per axis, the signal is clipped to the sensor range, and the resulting field is rotated into the sensor frame using the current orientation $\mathbf{R}_\text{sensor}$. Because heading is extracted from the horizontal projection of the measured magnetic field, any unmodeled inclination or anisotropic distortion introduces systematic and orientation-dependent yaw errors \cite{magnetometer}. To evaluate modeling fidelity, we compare the proposed model with a simplified fixed-reference magnetometer using a magnetometer-only tilt-compensated yaw estimator along a 3D AUV trajectory at 45° latitude. As shown in Fig.~\ref{fig:magnetometer_heading}, the simplified model, by neglecting geomagnetic inclination and explicit hard-/soft-iron effects, produces a systematic yaw bias and heading-dependent oscillatory errors caused by disturbance projection into the horizontal plane. In contrast, the proposed model incorporates latitude-dependent geomagnetic variation and explicit disturbance modeling prior to sensor-frame rotation, yielding bounded heading errors centered near zero with substantially reduced orientation sensitivity.

\section{Environmental Flow Modeling and Hydrodynamic Forces}

In addition to buoyancy, quadratic pressure drag, skin friction, and added mass, the simulator accounts for environmental flow effects and higher-order hydrodynamic forces relevant to realistic underwater operations. All hydrodynamic loads are evaluated directly on the vehicle’s triangulated surface mesh, enabling accurate treatment of partially submerged geometries and spatially varying flow fields. Fluid properties are sampled at the centroid of each submerged face, allowing the resulting forces and moments to reflect depth-dependent stratification and vertical shear. The velocity of the ambient fluid is provided by an ocean model that combines background currents with depth-dependent vertical shear. Vertical shear is represented as a smooth decay of current magnitude with depth, approximating a boundary-layer–like profile. Fluid density is evaluated using a \emph{depth-dependent} model in which density increases linearly within the pycnocline (approximately the first 1000~m) by up to 0.3\% above the surface value \cite{pycnocline}, capturing first-order effects of thermoclines and pressure-induced compression. This is implemented in the simulator via a simple linear ramp:

\begin{equation}
\rho(z) = \rho_0 + \Delta \rho_\mathrm{pyc} \, \frac{\min(z, z_\mathrm{pyc})}{z_\mathrm{pyc}},
\end{equation}

\noindent where $\rho_0$ is the surface density, $z_\mathrm{pyc}$ is the pycnocline depth, and $\Delta \rho_\mathrm{pyc}$ is the maximum pycnocline density increase. The resulting effective density $\rho_\mathrm{eff}$ is then used consistently across all hydrodynamic force and hydrostatic pressure calculations. While the rigid-body mass and collision geometry remain constant, the simulator scales the added mass, which accounts for fluid-induced resistance to translational acceleration, and the added inertia, which accounts for fluid-induced resistance to rotational acceleration, as functions of depth to capture pressure-induced compression of the displaced fluid volume and compliant structural components. Hydrostatic pressure is modeled as
\begin{equation}
p(d) = \int_0^d \rho(z) \, g \, dz,
\end{equation}
where $\rho(z)$ is the depth-dependent fluid density accounting for compression, $g$ is the gravitational acceleration, and $d$ denotes depth. A linear bulk modulus \cite{SHARQAWY2010354} approximation is then used to compute a compression factor
\begin{equation}
\kappa(d) = 1 - \frac{p(d)}{K},
\end{equation}
where $K$ is an effective bulk modulus. The depth-dependent added mass and added inertia are given by
\begin{equation}
\mathbf{M}_a(d) = \kappa(d) \, \mathbf{M}_a^{(0)}, \quad
\mathbf{I}_a(d) = \kappa(d) \, \mathbf{I}_a^{(0)},
\end{equation}
where $\mathbf{M}_a^{(0)}$ and $\mathbf{I}_a^{(0)}$ denote their nominal surface values corresponding to the undisturbed displaced fluid volume. This compression-aware formulation captures first-order deep-sea pressure effects on the vehicle dynamics, reflecting the reduction in displaced fluid volume, without introducing additional state variables or compromising real-time performance.  Lift is generated when pressure differences on a body produce a net force perpendicular to the relative flow velocity. For each submerged mesh face, the relative velocity at the face centroid is computed as
\begin{equation}
\mathbf{v}_c =
\mathbf{v}_{\text{fluid}}(\mathbf{r}_c)
-
\left(
\mathbf{v}
+
\boldsymbol{\omega} \times (\mathbf{r}_c - \mathbf{r}_{CG})
\right),
\end{equation}
where $\mathbf{v}_{\text{fluid}}(\mathbf{r}_c)$ is the local fluid velocity, $\mathbf{v}$ and $\boldsymbol{\omega}$ are the vehicle’s linear and angular velocities, and $\mathbf{r}_{CG}$ denotes the center of mass. The lift direction is defined to be orthogonal to both the face normal $\mathbf{n}$ and the relative velocity direction $\hat{\mathbf{v}}_c = \mathbf{v}_c / \|\mathbf{v}_c\|$,
\begin{equation}
\mathbf{d}_{\text{lift}} =
\frac{\mathbf{n} \times \hat{\mathbf{v}}_c}
{\|\mathbf{n} \times \hat{\mathbf{v}}_c\|}.
\end{equation}
The effective projected area is modulated by the local angle of attack according to
\begin{equation}
A_{\text{proj}} =
A \left(1 - \left| \mathbf{n} \cdot \hat{\mathbf{v}}_c \right| \right),
\end{equation}
ensuring that lift vanishes when the flow is aligned with the surface normal. The resulting lift force and corresponding moment contribution are given by
\begin{equation}
\mathbf{F}_{\text{lift}} =
\frac{1}{2}
\rho_{\text{eff}}
C_L
A_{\text{proj}}
\|\mathbf{v}_c\|^2
\mathbf{d}_{\text{lift}},
\,
\mathbf{T}_{\text{lift}} =
(\mathbf{r}_c - \mathbf{r}_{CG}) \times \mathbf{F}_{\text{lift}}.
\end{equation}

\begin{figure}[t]
    \centering
    \includegraphics[width=0.75\linewidth]{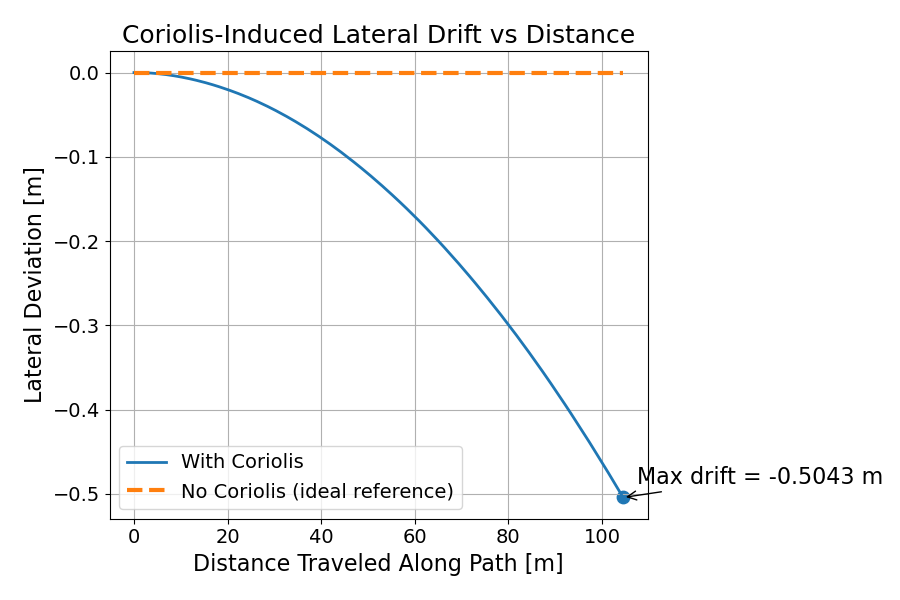}
\caption{Comparison of lateral displacement versus distance traveled with and without Coriolis effects. The deviation from the zero baseline indicates transverse deflection due to the Coriolis force.}
    \label{fig:coriolis_dirft}
    \vspace{-0.25cm}
\end{figure}

\begin{figure}[htbp]
    \centering
    \includegraphics[width=\linewidth]{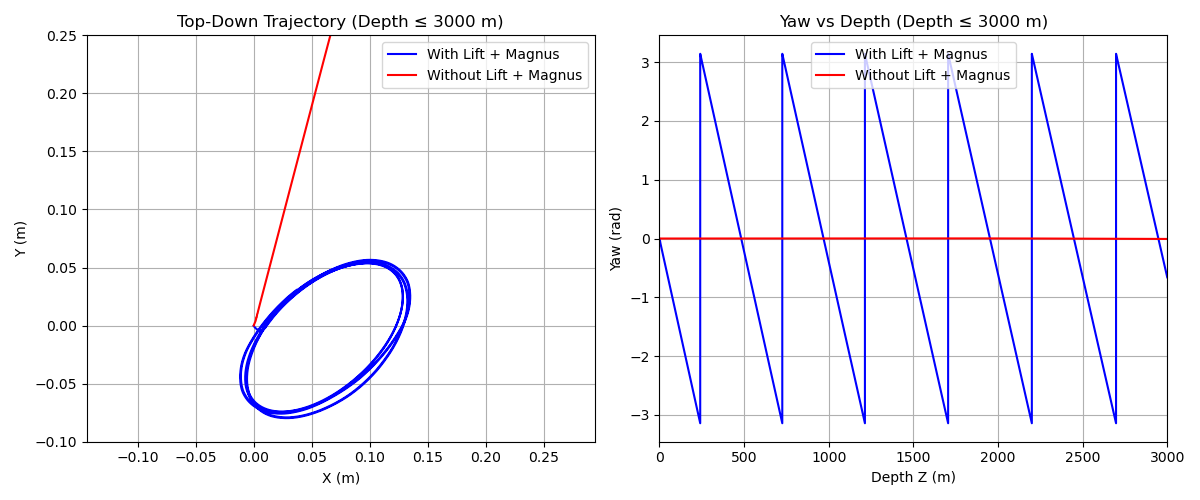}
     \caption{Top-down trajectory and yaw evolution of the lander with and without lift and Magnus effects up to 3000~m depth. 
(Left) XY trajectories. 
(Right) Yaw angle versus depth.}
    \label{fig:lander_lift_magnus}
        \vspace{-0.5cm}
\end{figure}

In addition to angle-of-attack lift, the simulator incorporates rotational lift effects arising from body rotation in a flowing fluid, commonly referred to as the Magnus effect. This force becomes significant for spinning or tumbling bodies and acts perpendicular to both the angular velocity and the local relative flow velocity. For each submerged face, the Magnus force contribution is computed as
\begin{equation}
\mathbf{F}_{\text{Magnus}} =
C_M
\rho_{\text{eff}}
A
\|\mathbf{v}_c\|
\left(
\boldsymbol{\omega} \times \mathbf{v}_c
\right),
\end{equation}
where $A$ is the face area and $C_M$ is a calibrated Magnus coefficient that implicitly incorporates characteristic length scaling. Although the Magnus effect is inherently a body-scale phenomenon, distributing the force across submerged mesh faces ensures numerical consistency with the surface-based force evaluation framework. The associated torque contribution is computed as
\begin{equation}
\mathbf{T}_{\text{Magnus}} =
(\mathbf{r}_c - \mathbf{r}_{CG}) \times \mathbf{F}_{\text{Magnus}}.
\end{equation}
This formulation captures stabilizing or destabilizing rotational effects in sheared or stratified flows without introducing additional state variables. Finally, Coriolis forces are included to account for inertial coupling between translational and rotational motion within the rigid-body dynamics formulation. These effects are modeled at the body level as
\begin{equation}
\mathbf{F}_{\text{coriolis}} = - m (\boldsymbol{\omega} \times \mathbf{v}),
\qquad
\mathbf{T}_{\text{coriolis}} = - (\mathbf{I} \boldsymbol{\omega}) \times \boldsymbol{\omega},
\end{equation}
where $m$ is the vehicle mass and $\mathbf{I}$ is its inertia tensor. Including these terms improves the fidelity of simulated dynamic responses during aggressive maneuvers and sustained rotational motion, complementing the surface-based hydrodynamic force model. Figure~\ref{fig:coriolis_dirft} illustrates the effect of Coriolis forces on the simulated vessel trajectory. Lateral deviation from an ideal straight path is shown as a function of distance traveled. The dashed line represents motion without Coriolis effects (zero deviation), while the measured curve shows the transverse drift that arises when Coriolis acceleration is included. The growing offset highlights the cumulative influence of rotational dynamics on long-distance motion prediction. Figure~\ref{fig:lander_lift_magnus} compares lander motion with and without lift and Magnus modeling across the considered depth range. When these effects are included, the lander exhibits a small but persistent lateral motion and associated yaw evolution arising from asymmetries in the interaction between body rotation and the surrounding flow, which generate lift forces and moments during descent. 
Such asymmetries are not captured by drag-only models but are expected for rotating bodies in a fluid and can accumulate over long-duration deep-water descents. The proposed lift and Magnus-effect models were validated against ANSYS CFD simulations and experimental measurements obtained from a free-fall deep-sea lander. 

\subsection{Seabed Interaction and Terramechanics Modeling}

In addition to fluid-mediated forces, the simulator accounts for direct mechanical interaction between the vehicle and the seabed, which frequently occurs during landing, resting, sampling, or benthic operation in deep-sea environments. Although such interactions are visually inconspicuous, they exert a dominant influence on vehicle dynamics, stability, and control, and are therefore critical for physically meaningful simulation. Seabed contact forces are evaluated using the same surface-based framework employed for hydrodynamic loads. For each mesh face intersecting the terrain, a penetration depth $z$ and local surface normal $\mathbf{n}_s$ are obtained from the terrain model. The normal contact pressure is computed using a lightweight pressure–sinkage relationship,
\begin{equation}
p_s(z) =
\left(
\frac{k_c}{b} + k_\phi
\right)
z^n,
\end{equation}
where $k_c$ and $k_\phi$ are the cohesive and frictional stiffness parameters in units of pressure [Pa], $n$ is the sinkage exponent, and $b=\sqrt{A}$ is the characteristic contact width for a face of area $A$. This formulation follows the classical terramechanics pressure–sinkage model of Bekker \cite{Bekker1969IntroductionTT} and has been widely applied to modeling ground-vehicle interactions \cite{article}. The resulting normal reaction force and torque contribution are given by
\begin{equation}
\mathbf{F}_{\text{soil}}^{(n)} = p_s(z) A \mathbf{n}_s,
\qquad
\mathbf{T}_{\text{soil}}^{(n)} =
(\mathbf{r}_c - \mathbf{r}_{CG}) \times \mathbf{F}_{\text{soil}}^{(n)}.
\end{equation}

\begin{figure}[t]
    \centering
    \includegraphics[width=1\linewidth]{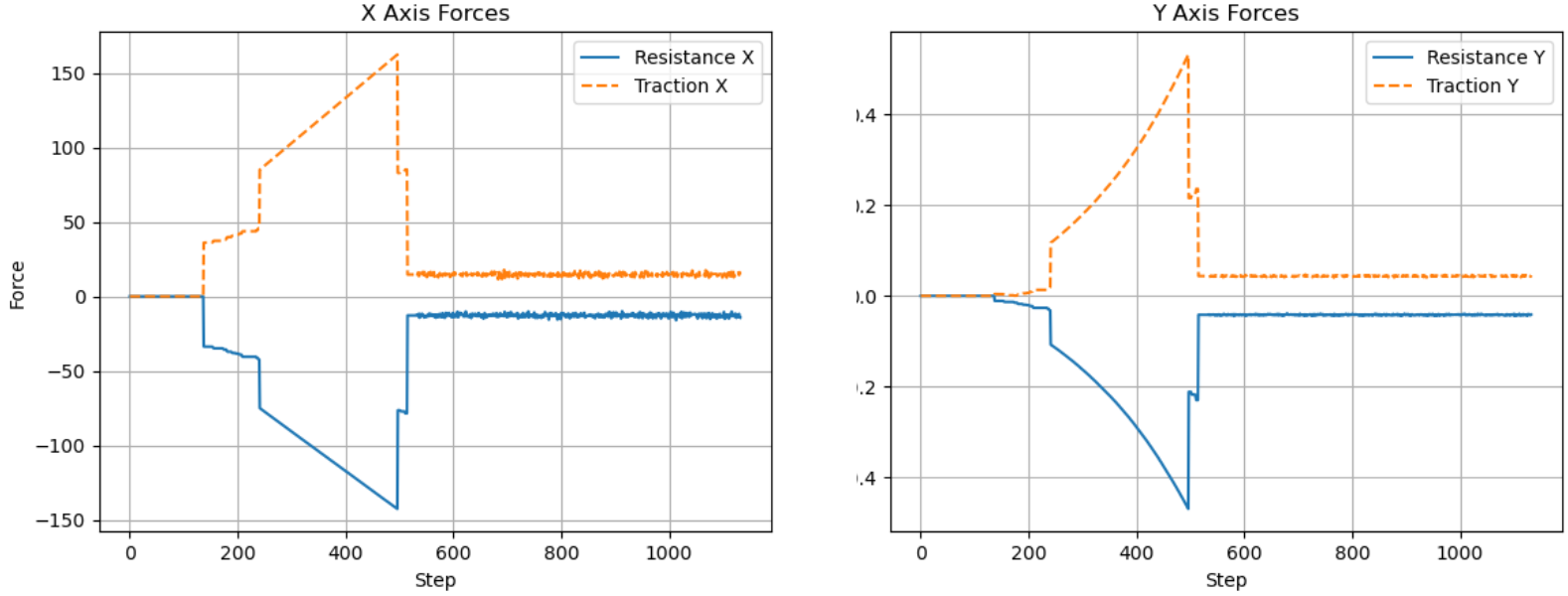}
    \caption{Soil–vehicle interaction forces. In the longitudinal direction (X), traction initially slightly exceeds resistance to generate forward acceleration; as velocity approaches steady state, the forces balance and stabilize. In the lateral direction (Y), forces remain symmetric due to the absence of motion.}
    \label{fig:terramech_forces}
    \vspace{-1.5em}
\end{figure}

Tangential resistance at each contact patch is limited by the soil's shear strength:
\begin{equation}
\tau_{\max} = c + p_s(z) \tan \phi,
\end{equation}
where $c$ is the soil cohesion, $\phi$ the internal friction angle, and $p_s(z)$ the local normal pressure. This sets the maximum tangential stress the soil can sustain; exceeding it causes plastic yielding. The tangential contact force opposes the local tangential velocity at the face centroid:
\begin{equation}
\mathbf{F}_{\text{soil}}^{(t)} =
- \tau_{\max} A
\frac{\mathbf{v}_t}{\|\mathbf{v}_t\|},
\end{equation}
where $\mathbf{v}_t$ is the tangential component of the contact point velocity. This represents the maximum tangential force; the actual force transmitted depends on slip, soil stiffness, and the contact patch size. To ensure numerical stability and physically dissipative behavior, a velocity-proportional normal damping term is additionally applied,
\begin{equation}
\mathbf{F}_{\text{soil}}^{(d)} =
- c_d A
\left(
\mathbf{v}_c \cdot \mathbf{n}_s
\right)
\mathbf{n}_s,
\end{equation}
where $c_d$ is a soil damping coefficient. This formulation captures sinkage, load-dependent shear resistance, and dissipative contact while remaining compatible with real-time simulation. Using the same per-face accumulation framework as hydrodynamics allows smooth transitions between free-flowing and grounded states without discrete contact modes. The resulting force evolution is shown in Fig.~\ref{fig:terramech_forces}, where longitudinal traction initially exceeds resistance during acceleration and stabilizes at steady velocity, consistent with 
the steady-state behavior of the tire-terrain interaction observed in 
finite element terramechanics studies \cite{Shoop2001FINITEEM}.

\begin{figure*}[]
    \centering
    \includegraphics[width=0.875\linewidth]{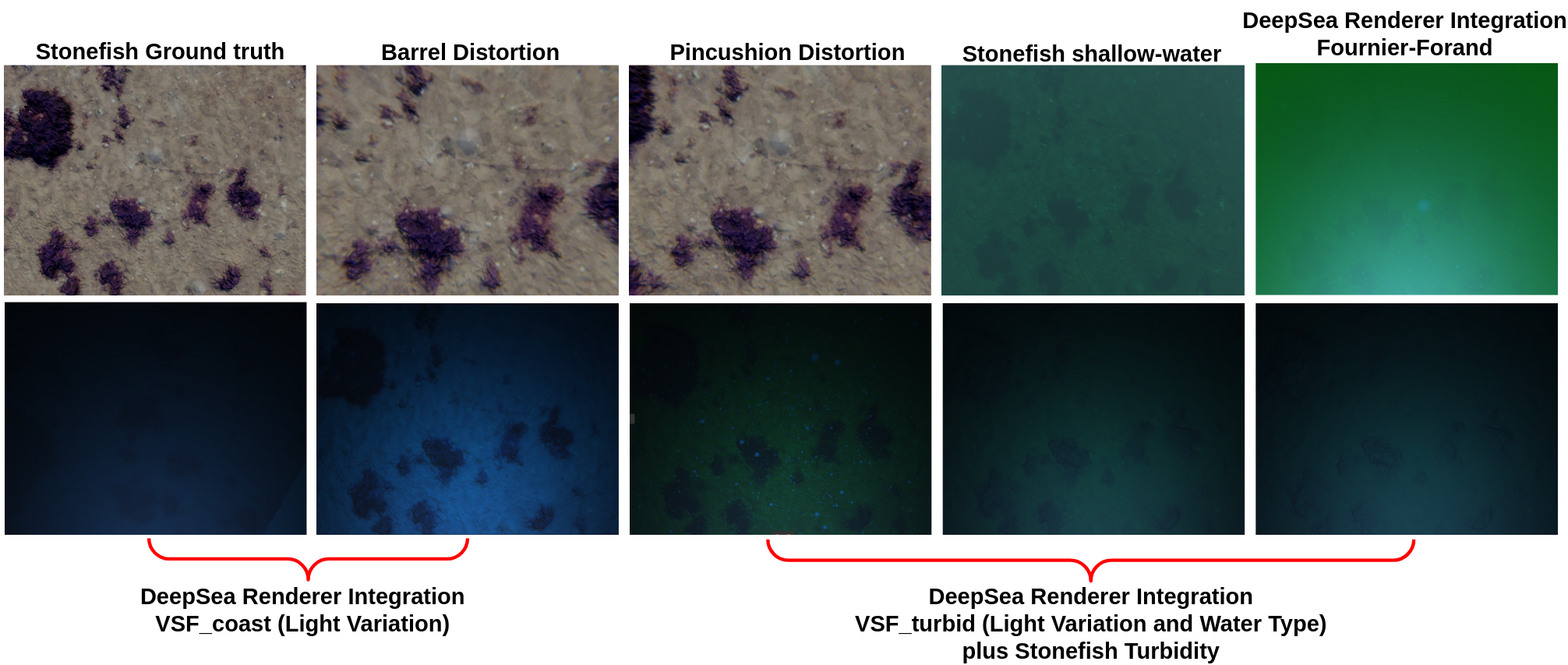}
\caption{Comparison of camera distortion and underwater rendering configurations. From left to right: original image, barrel distortion, pincushion distortion, a baseline shallow-water rendering produced by Stonefish, and DeepSeaRenderer outputs under varying illumination, scene geometry, depth, and scattering phase functions. }
    \vspace{-1.5em}
    \label{fig:Imagerendering}
\end{figure*}

\section{Underwater 3D Simulation and Rendering}

To support realistic underwater robotics experiments, Stonefish provides a modular framework for importing 3D models, simulating cameras, and rendering deep-sea environments, as described in the following subsections.

\subsection{GLTF/GLB Model Import and Optimization}

\begin{figure*}[h]
    \centering
    \includegraphics[width=0.9\linewidth]{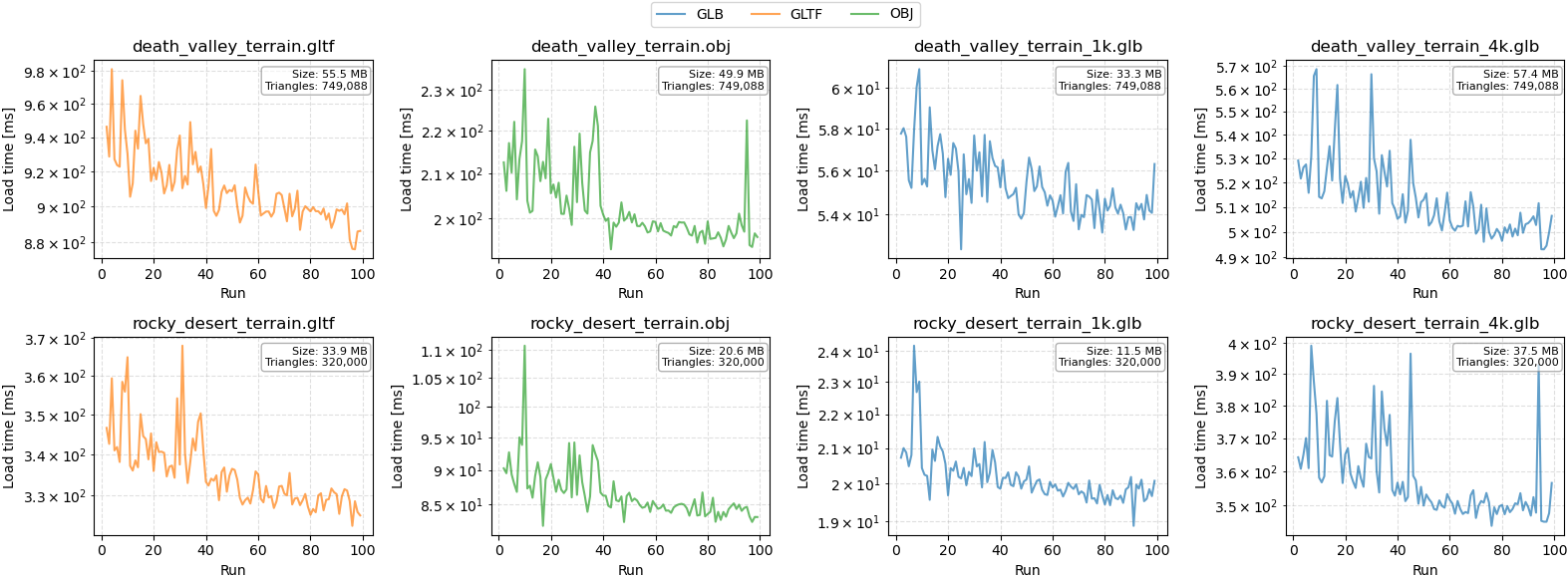}
\caption{Loading times for the same 3D object with different formats (GLTF, GLB, OBJ) and quality levels. The first row shows the object with a lower triangle count, and the second row shows the higher triangle count version. Decreases in load time for later runs are observed due to memory and file system caching, which can make subsequent loads faster.}
    \vspace{-0.5cm}
    \label{fig:model_loading_times}
\end{figure*}

The simulator supports importing 3D models in GLTF \footnote{\url{https://www.khronos.org/gltf/}} and GLB formats. Currently, the implementation focuses on mesh loading. Depending on the file extension, the loader parses either a binary GLB or ASCII GLTF file. It then traverses the scene hierarchy, extracting mesh primitives composed of vertices, normals, and optional UV coordinates, which are assembled into internal mesh structures and scaled to match simulator units. Large meshes are handled efficiently in a single pass to reduce memory allocations and improve cache coherence, enabling rapid loading even for models containing millions of triangles. Thanks to these optimizations, the simulator can efficiently load large GLTF/GLB models; full PBR material support is planned for future work.
Figure~\ref{fig:model_loading_times} illustrates the loading times for various model sizes, demonstrating that even complex scenes can be imported in a matter of seconds, enabling fast iteration and testing within the simulation environment.

\subsection{Parametric Underwater Camera Modelling}

Accurate modelling of camera projection and image distortion is important for evaluating vision-based algorithms in simulation. Stonefish provides a configurable camera framework for generating underwater imagery from GLTF/GLB scenes. It supports rectilinear, fisheye, and wide-angle configurations with adjustable field of view, intrinsic parameters, and image-plane distortion. Lens distortion is represented using radial coefficients $k_1$, $k_2$, and $k_3$, together with optional tangential coefficients $p_1$ and $p_2$. This parameterisation follows the OpenCV radial--tangential model, allowing simulated cameras to be configured using real calibration data.
For normalised image coordinates $(x,y)$, let $r^2=x^2+y^2$ and define the radial scale factor as
\begin{equation}
s_r = 1+k_1r^2+k_2r^4+k_3r^6 .
\end{equation}
The distorted coordinates are then
\begin{equation}
\begin{aligned}
x_d &= xs_r+2p_1xy+p_2(r^2+2x^2),\\
y_d &= ys_r+p_1(r^2+2y^2)+2p_2xy .
\end{aligned}
\end{equation}

The corresponding pixel coordinates are obtained as $u=f_xx_d+c_x$ and $v=f_yy_d+c_y$, where $f_x$ and $f_y$ are the focal lengths and $(c_x,c_y)$ is the principal point. These transformations are applied within the OpenGL rendering pipeline, enabling real-time generation of calibrated barrel, pincushion, and tangential distortion, as illustrated in Fig.~\ref{fig:Imagerendering}.

Stonefish does not explicitly simulate light propagation through water, housing glass, and air, and currently does not support flat- or dome-port refraction, including depth-dependent and non-central projection effects. The formulation should therefore be interpreted as an efficient calibrated approximation for real-time underwater vision benchmarking rather than a complete physical model of underwater camera housings.

\subsection{Deep-Sea Rendering Integration}

Stonefish provides a lightweight rendering pipeline tightly coupled with
its physics engine, enabling simulation of underwater vehicles, ocean
surface effects, and underwater scenes. Its native renderer includes
wavelength-dependent absorption and scattering models based on Jerlov
water types~\cite{Jerlov01011951}, which are well suited for shallow-water
and coastal scenarios. Deep-sea environments, however, differ
substantially: natural light is nearly absent, artificial illumination
dominates image formation, and imagery is strongly affected by attenuation,
backscatter, and illumination-dependent colour shifts. To better approximate these conditions, we connect Stonefish with the
DeepSea renderer~\cite{song2021deep,song2022virtually} at the ROS
image-pipeline level. Stonefish publishes rendered RGB and depth images,
which are processed by DeepSea to synthesise effects such as co-moving
artificial illumination, scattering, and radiometric attenuation. The
processed images are then republished as enhanced camera streams, allowing
existing perception and navigation algorithms to operate on more realistic
deep-sea imagery while preserving Stonefish's vehicle dynamics, sensor
timing, and scene geometry. This modular interface links Stonefish’s physics simulation with DeepSea’s radiometric image formation and supports future native integration.

\section{Conclusion}
This work advocates for a shift in underwater simulation toward physically and sensor-grounded modeling that reflects the operational realities of deep-ocean robotics. Rather than treating visual fidelity as the primary indicator of realism, we emphasize modeling the forces, measurements, and environmental conditions that directly influence robotic behavior and autonomy, particularly in deep and low-visibility settings. The benefits of this approach extend beyond traditional AUV and ROV simulation. Institutions developing crawlers, landers, surface vehicles, and hybrid platforms can benefit from a unified simulation framework that supports diverse vehicle types and mission profiles, especially where priorities include navigation robustness, control stability, perception under low signal-to-noise conditions, and long-duration autonomy, becoming a more effective tool for system development, integration, and validation. This physics- and sensor-centric emphasis also provides a foundation for future deep-sea digital twin applications, building on existing underwater digital twin concepts such as \cite{adetunji2024digitaltwinssurfaceenhancing}, as vehicle-specific parameters, operational data, and environmental models are incorporated. While significant challenges remain, this work represents a step toward more practical and broadly applicable underwater simulation, and we hope it supports ongoing efforts to advance the field.

\vspace{-1em}
\bibliographystyle{IEEEtran}
\bibliography{biblio}

@inproceedings{song2021deep,
  title={Deep Sea Robotic Imaging Simulator},
  author={Song, Yifan and Nakath, David and She, Mengkun and Elibol, Furkan and Koser, Kevin},
  booktitle={Pattern Recognition. ICPR International Workshops and Challenges},
  year={2021},
  publisher={Springer},
  pages={375--389},
  doi={https://doi.org/10.1007/978-3-030-68790-8_29}
}

@article{Jerlov01011951,
author = {N. G. Jerlov},
title = {Optical Measurement of Particle Distribution in the Sea},
journal = {Tellus},
volume = {3},
number = {3},
pages = {122--128},
year = {1951},
publisher = {Taylor \& Francis},

}

@article{role_of,
author = {Liu, C. and Negrut, Dan},
year = {2021},
month = {05},
pages = {},
title = {The Role of Physics-Based Simulators in Robotics},
journal = {Annual Review of Control, Robotics, and Autonomous Systems},
doi = {10.1146/annurev-control-072220-093055}
}

@article{aldhaheri2025underwaterroboticsimulatorsreview,
  title={Underwater Robotic Simulators Review for Autonomous System Development},
  author={Sara Aldhaheri and Yang Hu and Yongchang Xie and Peng Wu and Dimitrios Kanoulas and Yuanchang Liu},
  journal={OCEANS 2025 Brest},
  year={2025},
  pages={01-10},
  urlnot={https://api.semanticscholar.org/CorpusID:277628070}
}

@ARTICLE{712192,
  author={Sutton, R.S. and Barto, A.G.},
  journal={IEEE Transactions on Neural Networks}, 
  title={Reinforcement Learning: An Introduction}, 
  year={1998},
  doi={10.1109/TNN.1998.712192}}

@INPROCEEDINGS{adetunji2024digitaltwinssurfaceenhancing,
  author={Adetunji, Favour O. and Ellis, Niamh and Koskinopoulou, Maria and Carlucho, Ignacio and Petillot, Yvan R.},
  booktitle={OCEANS 2024 - Singapore}, 
  title={Digital Twins Below the Surface: Enhancing Underwater Teleoperation}, 
  year={2024},
  volume={},
  number={},
  pages={1-8},
  doi={10.1109/OCEANS51537.2024.10682270}}

@ARTICLE{5570702,
  author={Barnes, James A. and Chi, Andrew R. and Cutler, Leonard S. and Healey, Daniel J. and Leeson, David B. and McGunigal, Thomas E. and Mullen, James A. and Smith, Warren L. and Sydnor, Richard L. and Vessot, Robert F. C. and Winkler, Gernot M. R.},
  journal={IEEE Transactions on Instrumentation and Measurement}, 
  title={Characterization of Frequency Stability}, 
  year={1971},}

@misc{amer2023unavsimvisuallyrealisticunderwater,
      title={UNav-Sim: A Visually Realistic Underwater Robotics Simulator and Synthetic Data-generation Framework}, 
      author={Abdelhakim Amer and Olaya Álvarez-Tuñón and Halil Ibrahim Ugurlu and Jonas le Fevre Sejersen and Yury Brodskiy and Erdal Kayacan},
      year={2023},
      eprint={2310.11927},
      archivePrefix={arXiv},
      primaryClass={cs.RO},
      urlnot={https://arxiv.org/abs/2310.11927}, 
}

@inproceedings{Shoop2001FINITEEM,
  title={FINITE ELEMENT MODELING OF TIRE-TERRAIN INTERACTION},
  author={Sally A. Shoop},
  year={2001},
}

@article{magnetometer,
author = {Feng, Ye and Shi, Feng and Lai, Yizong and Zhou, Xinjie and Li, Kuo},
year = {2020},
month = {10},
pages = {},
title = {Heading angle estimation using rotating magnetometer for mobile robots under environmental magnetic disturbances},
volume = {13},
journal = {Intelligent Service Robotics},
doi = {10.1007/s11370-020-00334-7}
}

@inproceedings{chu2025marinegymhighperformancereinforcementlearning,
  title={MarineGym: a high-performance reinforcement learning platform for underwater robotics},
  author={Chu, Shuguang and Huang, Zebin and Li, Yutong and Lin, Mingwei and Li, Dejun and Carlucho, Ignacio and Petillot, Yvan R and Yang, Canjun},
  booktitle={2025 IEEE/RSJ International Conference on Intelligent Robots and Systems (IROS)},
  pages_not={17146--17153},
  year={2025},
  organization={IEEE}
}

@Article{app13020680,
AUTHOR = {Yoon, Jaemin and Son, Bukun and Lee, Dongjun},
TITLE = {Comparative Study of Physics Engines for Robot Simulation with Mechanical Interaction},
JOURNAL = {Applied Sciences},
VOLUME = {13},
YEAR = {2023},
NUMBER = {2},
ARTICLE-NUMBER = {680},
urlnot = {https://www.mdpi.com/2076-3417/13/2/680},
ISSN = {2076-3417},
DOI = {10.3390/app13020680}
}

@article{article,
author = {Grazioso, Andrea and di Maria, Enrico and Giannoccaro, Nicola and Ishii, Kazuo},
year = {2022},
month = {11},
pages = {1117},
title = {Multibody Modeling of a New Wheel/Track Reconfigurable Locomotion System for a Small Farming Vehicle},
volume = {10},
journal = {Machines},
doi = {10.3390/machines10121117}
}

@inproceedings{Bekker1969IntroductionTT,
  title={Introduction to Terrain-Vehicle Systems},
  author={Mieczysław G. Bekker},
  year={1969},
  urlnot={https://api.semanticscholar.org/CorpusID:131292053}
}

@misc{tu2019gapmodelbasedmodelfreemethods,
      title={The Gap Between Model-Based and Model-Free Methods on the Linear Quadratic Regulator: An Asymptotic Viewpoint}, 
      author={Stephen Tu and Benjamin Recht},
      year={2019},
      eprint={1812.03565},
      archivePrefix={arXiv},
      primaryClass={cs.LG},
      urlnot={https://arxiv.org/abs/1812.03565}, 
}

@article{osti_2282016,
  author       = {Karniadakis, George Em and Kevrekidis, Ioannis G. and Lu, Lu and Perdikaris, Paris and Wang, Sifan and Yang, Liu},
  title        = {Physics-informed machine learning},
  doi          = {10.1038/s42254-021-00314-5},
  journal      = {Nature Reviews Physics},
  issn         = {ISSN 2522-5820},
  number       = {6},
  volume       = {3},
  place        = {United States},
  publisher    = {Springer Nature},
  year         = {2021},
  month        = {05}}

@ARTICLE{10638434,
  author={Potokar, Easton and Lay, Kalliyan and Norman, Kalin and Benham, Derek and Ashford, Spencer and Peirce, Randy and Neilsen, Tracianne B. and Kaess, Michael and Mangelson, Joshua G.},
  journal={IEEE Journal of Oceanic Engineering}, 
  title={HoloOcean: A Full-Featured Marine Robotics Simulator for Perception and Autonomy}, 
  year={2024},}

@misc{zhang2022daveaquaticvirtualenvironment,
      title={DAVE Aquatic Virtual Environment: Toward a General Underwater Robotics Simulator}, 
      author={Mabel M. Zhang and Woen-Sug Choi and Jessica Herman and Duane Davis and Carson Vogt and Michael McCarrin and Yadunund Vijay and Dharini Dutia and William Lew and Steven Peters and Brian Bingham},
      year={2022},
      eprint={2209.02862},
      archivePrefix={arXiv},
      primaryClass={cs.RO},
      urlnot={https://arxiv.org/abs/2209.02862}, 
}

@misc{song2025oceansimgpuacceleratedunderwaterrobot,
      title={OceanSim: A GPU-Accelerated Underwater Robot Perception Simulation Framework}, 
      author={Jingyu Song and Haoyu Ma and Onur Bagoren and Advaith V. Sethuraman and Yiting Zhang and Katherine A. Skinner},
      year={2025},
      eprint={2503.01074},
      archivePrefix={arXiv},
      primaryClass={cs.RO}, 
}

@article{song2022virtually,
  title={Virtually throwing benchmarks into the ocean for deep sea photogrammetry and image processing evaluation},
  author={Song, Yifan and She, Mengkun and K{\"o}ser, Kevin},
  journal={ISPRS Annals of the Photogrammetry, Remote Sensing and Spatial Information Sciences},
  volume={V-4-2022},
  pages={353--360},
  year={2022},
  doi={https://doi.org/10.1007/978-3-030-68790-8_29}
}

@inproceedings{stonefish,
   author = "Cie{\'s}lak, PatryHk",
   booktitle = "OCEANS 2019 - Marseille",
   title = "Stonefish: An Advanced Open-Source Simulation Tool Designed for Marine Robotics, With a ROS Interface",
   month = "jun",
   year = "2019",
   doi = "10.1109/OCEANSE.2019.8867434"
}

@misc{grimaldi2025stonefishsupportingmachinelearning,
      title={Stonefish: Supporting Machine Learning Research in Marine Robotics}, 
      author={Michele Grimaldi and Patryk Cieslak and Eduardo Ochoa and Vibhav Bharti and Hayat Rajani and Ignacio Carlucho and Maria Koskinopoulou and Yvan R. Petillot and Nuno Gracias},
      year={2025},
      eprint={2502.11887},
      archivePrefix={arXiv},
      primaryClass={cs.RO},
      urlnot={https://arxiv.org/abs/2502.11887}, 
}

@article{SHARQAWY2010354,
title = {Thermophysical properties of seawater: a review of existing correlations and data},
journal = {Desalination and Water Treatment},
volume = {16},
number = {1},
pages = {354-380},
year = {2010},
issn = {1944-3986},
doi = {https://doi.org/10.5004/dwt.2010.1079},
author = {Mostafa H. Sharqawy and John H. Lienhard and Syed M. Zubair}
}

@book{pycnocline,
author = {Stewart, Robert},
year = {2008},
month = {01},
pages = {353},
title = {Introduction To Physical Oceanography},
volume = {65},
isbn = {0132381559},
journal = {American Journal of Physics},
doi = {10.1119/1.18716}
}

\end{document}